\documentclass[10pt,twocolumn,letterpaper]{article}

\usepackage[pagenumbers]{wacv}

\definecolor{wacvblue}{rgb}{0.21,0.49,0.74}
\usepackage[pagebackref,breaklinks,colorlinks,allcolors=wacvblue]{hyperref}

\usepackage{graphicx}
\usepackage{amsmath}
\usepackage{amssymb}
\usepackage{booktabs}
\usepackage{multirow}
\usepackage{xcolor}
\usepackage{pifont}
\usepackage{algorithm}
\usepackage{algpseudocode}

\newcommand{\cmark}{\ding{51}}
\newcommand{\xmark}{\ding{55}}

\def\I{\mathbf{I}}
\def\A{\mathbf{A}}
\def\M{\mathbf{M}}

\def\R{\mathbb{R}}

\begin{document}

\title{Segmenting Collision Sound Sources in Egocentric Videos}

\author{Kranti Kumar Parida\textsuperscript{1} \quad Omar Emara\textsuperscript{2} \quad Hazel Doughty\textsuperscript{3} \quad Dima Damen\textsuperscript{2}\\
\textsuperscript{1}Samsung R\&D Institute India – Bangalore \quad \textsuperscript{2}University of Bristol \quad \textsuperscript{3}Leiden University \\
\url{https://krantiparida.github.io/projects/cs3.html}
\thanks{This work was done when Kranti was at University of Bristol.}
}
\maketitle
\vspace{-2 em}

\vspace{-0.5em}
\begin{abstract}
Humans excel at multisensory perception and can often recognise object properties from the sound of their interactions. Inspired by this, we propose the novel task of \underline{C}ollision \underline{S}ound \underline{S}ource \underline{S}egmentation~(CS3), where we aim to segment the objects responsible for a collision sound in visual input (\ie video frames from the collision clip), conditioned on the audio. This task presents unique challenges. Unlike isolated sound events, a collision sound arises from interactions between two objects, and the acoustic signature of the collision depends on both. We focus on egocentric video, where sounds are often clear, but the visual scene is cluttered, objects are small, and interactions are brief. 

To address these challenges, we propose a weakly-supervised method 
for audio-conditioned segmentation, utilising
foundation models (CLIP and SAM2).
We also incorporate egocentric cues, \ie objects in hands, to find acting objects that can potentially be collision sound sources. Our approach outperforms competitive baselines by $3\times$ and $4.7\times$ in mIoU on two benchmarks we introduce for the CS3 task: EPIC-CS3 and Ego4D-CS3.
\end{abstract}    
\vspace{-0.8em}
\section{Introduction}
\vspace{-0.3em}

\label{sec:intro}
Humans effortlessly interpret the world around them using both visual and auditory cues. Sounds from physical interactions, 
such as a spoon clattering against a bowl or a hammer striking a nail, carry rich information about these actions or events. 
Despite their richness and frequency in everyday interactions, these \textit{collision sounds} are rarely studied in audio-visual modelling.
Such \textit{collision sounds} are especially informative in egocentric video where the interaction sounds are clearly captured by wearable microphones, even in cluttered or visually ambiguous scenes. Auditory cues reveal the nature of interacting objects, enabling quick and intuitive reasoning for humans and embodied agents alike. %

\begin{figure}
    \centering
    \includegraphics[width=0.99\linewidth]{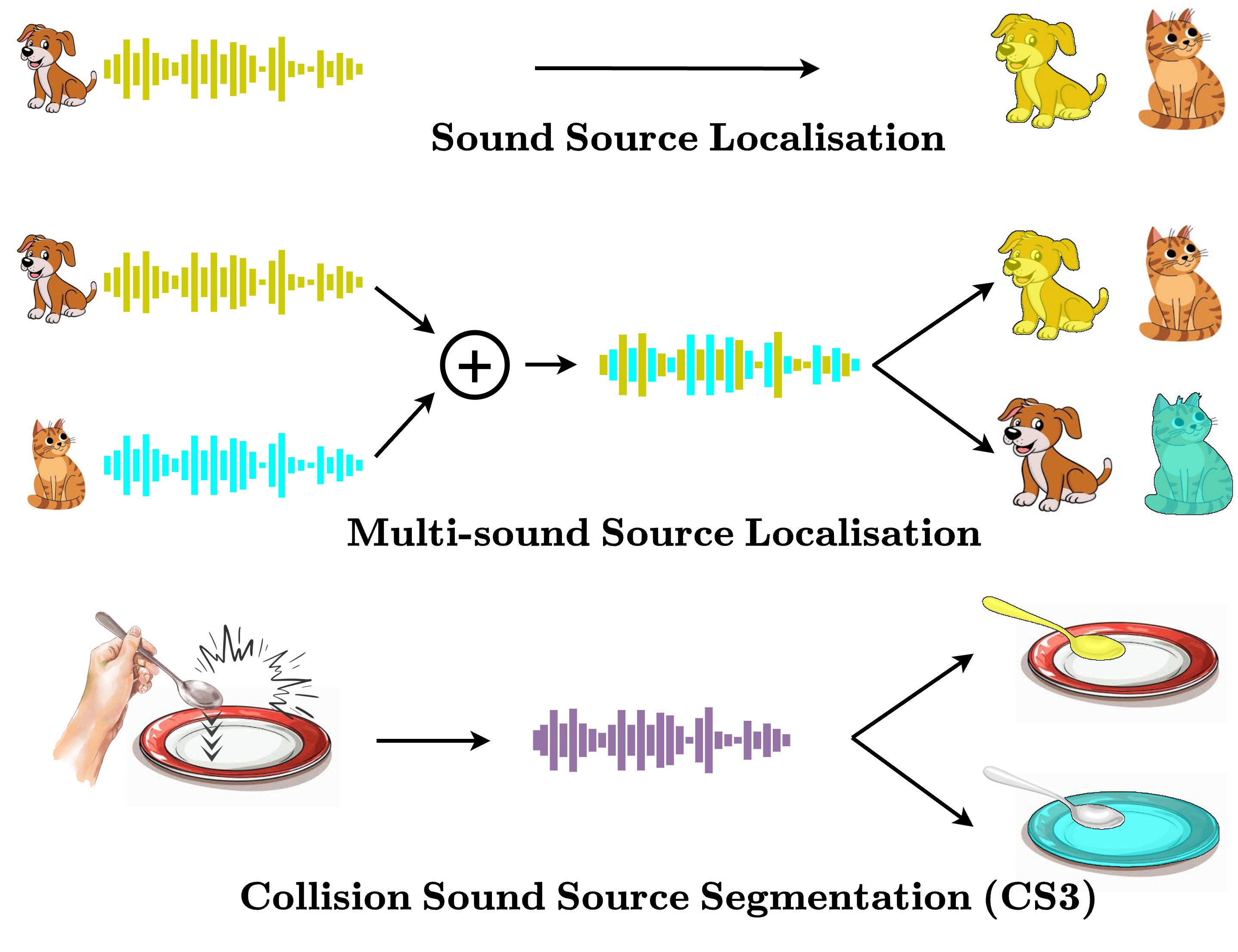}
    \vspace{-1.2 em}
    \caption{
    \textbf{\textbf{Task Definitions.} (Top)} Sound source localisation segments one object given its sound. \textbf{(Middle)} Multi-sound source localisation segments multiple objects from a mixture of their individually distinct sounds. \textbf{(Bottom)} We introduce a novel task, CS3, to segment the sources of \textit{collision sound} by identifying the objects involved in the interaction, based on the impact sound.
}
    \label{fig:teaser}
    \vspace{-1.5 em}
\end{figure}

Prior works localising sound sources~\cite{senocak2018learning, chen2021localizing, senocak2023sound, hu2022mix, mo2023audio} have made progress in linking sounds to objects. However, these efforts focus on isolated, clearly audible objects, like a dog barking or a car passing, where sound maps directly to a single object (Fig.~\ref{fig:teaser} top). Multi-sound source localisation extends this by localising multiple objects producing sounds at the same time, however each has a distinct sound (\eg a dog barking and cat meowing, Fig.~\ref{fig:teaser} middle). 
While the recoded audio is the mixture of both sounds, each object still produces an independent sound that can be separated from the mixed audio. Moreover, datasets for these tasks often have minimal visual clutter or noise~\cite{senocak2018learning, chen2021localizing,zhou2022audio}. %

In contrast, we introduce the task of \underline{C}ollision \underline{S}ound \underline{S}ource \underline{S}egmentation~(CS3) (Fig.~\ref{fig:teaser}, bottom), where given a collision sound and a corresponding image the goal is to provide pixel segmentations of the object(s) producing this collision sound.
We focus on egocentric audio-visual clips~\cite{damen2018scaling,Grauman_2022_CVPR} where collision sounds are prominent~\cite{huh2023epic}.

\textit{Collision-induced sounds} pose fundamentally different challenges. First, these sounds are not produced by individual objects but by interactions between two objects or parts of the same object. The collision sound is defined by the characteristics of both objects as well as the type and force of interaction; and is not a linear combination of two sounds. For example, a metal spoon sounds different hitting a ceramic plate than hitting a plastic container. Second, collisions in egocentric videos often occur in cluttered environments with frequent occlusions. 
Large-scale segmentation annotations are difficult to obtain in these settings. 

We thus design a weakly supervised method that requires only collision clips and can segment collision sound sources without segmentation annotations. Our approach uses a CLIP-based audio-conditioned segmentation model. Given most egocentric collisions involve hand-object interactions, we additionally utilise this prior with a hand-object interaction model that locates in-hand objects. We then verify the colliding objects to segment either two objects causing the collision, or a single object with colliding parts. 

Our contributions are threefold. First, we introduce the task of collision sound source segmentation (CS3), which aims to identify objects responsible for a collision using an audio-visual clip of the impact. Second, we curate two challenging CS3 datasets from large-scale egocentric datasets EPIC-Kitchens~\cite{damen2018scaling} and Ego4D~\cite{Grauman_2022_CVPR}. %
Compared to existing sound source localisation datasets \cite{senocak2018learning, chen2021localizing, zhou2022audio}, our scenes are more cluttered, objects are smaller ($\sim$10\% of the image), and sounds arise from collisions rather than isolated sources. Third, we propose a method combining audio-conditioned CLIP-based segmentation with hand-object reasoning and show it significantly outperforms baselines on both datasets.

\vspace{-0.3em}
\section{Related Works}
\vspace{-0.2em}
\label{sec:related_works}
Audio-visual learning~\cite{arandjelovic2017look, arandjelovic2018objects, korbar2018cooperative, morgado2021audio, owens2018audio, morgado2020learning, yang2020telling} exploits correlation between audio and visual modalities to learn aligned representations for various downstream tasks. In this work, we focus on localising objects involved in collision, and therefore review prior work on general sound source localisation~\cite{senocak2018learning, chen2021localizing, mo2022localizing, mo2022closer, kim2025improving, huang2023egocentric, park2024can, hu2020discriminative, qian2020multiple, hu2022mix, mo2023audio, kim2024learning, mahmud2024t, um2025object} as well as audio-visual methods specifically addressing collision or impact sounds~\cite{arnab2016joint, owens2016visually, zhang2017generative, sterling2018isnn, shi2021glavnet, gao2021objectfolder, gao2022objectfolder, clarke2023realimpact, clarke2025x, bagad2025sound, wilson2019analyzing}.

\noindent \textbf{Sound Source Localisation} identifies the image region responsible for a given sound~\cite{senocak2018learning, chen2021localizing, mo2022localizing, mo2022closer, senocak2023sound, fedorishin2023hear, kim2025improving, huang2023egocentric, park2024can}. A common strategy is to learn audio-visual correspondence via a contrastive loss on paired image-audio data \cite{senocak2018learning}. Variants enhance performance by incorporating pre-trained object detectors~\cite{mo2022localizing}, suppressing irrelevant image regions~\cite{mo2022closer}, using hard negatives \cite{chen2021localizing}, increasing positive pairs \cite{senocak2023sound}, adding optical flow \cite{fedorishin2023hear}, or leveraging slot-attention \cite{kim2025improving}. While most methods target object sounds, \cite{huang2023egocentric} addressed action sounds in videos using geometric pooling of visual features and synthetic audio mixtures.

Another direction adapts pre-trained CLIP-based vision-language models~\cite{radford2021learning} for sound source localisation via audio-visual correspondence. The method~\cite{park2024can} combines CLIP with supervised segmentation to improve localisation performance after adapting to audio inputs. Our approach is inspired by these works, following the same principle of weak supervision. Different to existing methods which  assume a single sound source per image, we aim to localise collision sounds that often arise from multiple objects.

\noindent\textbf{Multi-sound Source Localisation} aims to spatially localise multiple sounding objects in a scene, each corresponding to distinct a audio cue~\cite{hu2020discriminative, qian2020multiple, hu2022mix, mo2023audio, kim2024learning, mahmud2024t, um2025object}. Hu \etal \cite{hu2020discriminative} learn a dictionary of sounding objects from single-source audio-visual pairs, and use this to localise multiple sources in mixtures. Qian \etal~\cite{qian2020multiple} adopt a coarse-to-fine strategy with class labels. Mo~\etal~\cite{mo2023audio} use semantic labels for joint classification and localisation. Hu \etal~\cite{hu2022mix} apply a cycle consistency loss to match audio sources to objects. Kim \etal~\cite{kim2024learning} localise an unknown number of sources using optical flow alongside RGB images. Other works~\cite{mahmud2024t, um2025object} use auxiliary text information to aid localisation. While related, our task differs fundamentally from this line of work. In multi-object sound source localisation, audio is a mix of independent sources; in our case, sounds emerge from object interactions and cannot be attributed to either object alone. This requires reasoning about interaction-induced audio, not just source separation. 

\begin{figure*}
    \centering
    \includegraphics[width=0.99\linewidth]{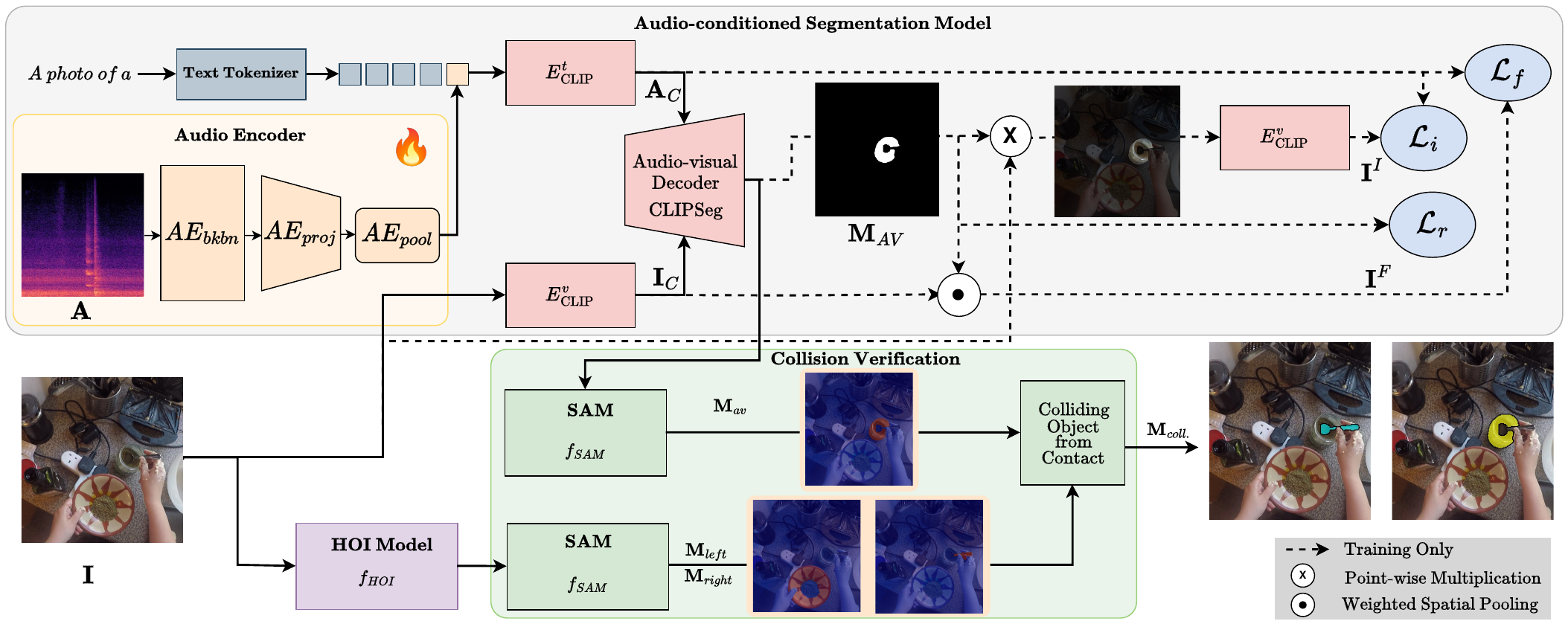}
    \vspace{-0.5em}
    \caption{\textbf{Proposed Architecture} Our architecture consists of three main components: (1) audio-conditioned segmentation, (2) hand-object interaction (HOI) and (3) collision verification. The audio-conditioned segmentation model takes an image ($\I$) and its corresponding audio ($\A$) to produce conditioning signals $\I_C$ and $\A_C$. The audio is first encoded into a representation aligned with the text token space, which is used alongside visual features to guide the localisation of sound-producing regions. The model is trained with image-level ($\mathcal{L}_{i}$), feature-level ($\mathcal{L}_{f}$), area regaularisation ($\mathcal{L}_{r}$) losses. The HOI model provides bounding boxes for in-hand left and right objects when present. The collision verification module uses SAM to extract object masks for audio-conditioned segmentation mask $\M_{av}$ and in-hand objects $\M_{\textrm{\textit{left}}}$ and $\M_{\textrm{\textit{right}}}$. A contact-based strategy is then applied to estimate the segmentations for collision sound sources, $\M_{coll.}$.
    }
    \vspace{-0.3em}
    \label{fig:approach}
\end{figure*}

\noindent\textbf{Audio-Visual Collision Datasets}. 
Several audio-visual datasets involving collision sounds have been introduced \cite{arnab2016joint, owens2016visually, zhang2017generative, sterling2018isnn, shi2021glavnet, gao2021objectfolder, gao2022objectfolder, clarke2023realimpact, clarke2025x, bagad2025sound, wilson2019analyzing}. Early work \cite{arnab2016joint, owens2016visually} recorded real objects being struck with a drumstick or knuckle. Subsequent datasets introduced synthesised impact sounds~\cite{zhang2017generative}, later enhanced with real recordings and impact information~\cite{sterling2018isnn, shi2021glavnet}. 
Others added tactile data for synthetic objects~\cite{gao2021objectfolder, gao2022objectfolder}, or multichannel impact sounds recorded from automated setups in controlled environments~\cite{clarke2023realimpact}. X-Capture~\cite{clarke2025x} relaxed these constraints, collecting audio, visual, and tactile data in natural settings with a portable device. 
Beyond collisions, a few datasets capture liquid pouring sounds \cite{bagad2025sound, wilson2019analyzing}, enabling inference of physical properties such as container shape and liquid mass.

However, most existing datasets are either simulated~\cite{zhang2017generative, sterling2018isnn, shi2021glavnet, gao2021objectfolder, gao2022objectfolder} or recorded in controlled settings~\cite{clarke2023realimpact} with a fixed striking object~\cite{arnab2016joint, owens2016visually}. By standardising one object across the dataset, these works focus on the \textit{other} object in the collision. 
Such datasets are unsuitable for our task as they contain a single object in view struck with a standard \textit{hitting} tool. While Arnab \etal~\cite{arnab2016joint} provide segmentation masks, their dataset is limited to 600 samples, with collisions always involving the same knuckle. 
This is why prior \textit{collision sound} datasets are not used for localisation and are primarily used for cross-modal alignment~\cite{owens2016visually, gao2021objectfolder, gao2022objectfolder, clarke2023realimpact} and material/shape classification~\cite{owens2016visually, zhang2017generative, sterling2018isnn, shi2021glavnet}. 

In contrast, we aim to segment collision sound sources from natural daily activity videos, where collisions occur amid clutter, occlusion and background noise. To our knowledge, we are the first to introduce the task of collision sound source segmentation (CS3) in unconstrained settings.

\vspace{-0.2em}
\section{CS3: Problem Definition}
\vspace{-0.3em}
We define our task, Collision Sound Source Segmentation~(CS3) as follows:
given an audio-visual clip containing a single collision sound, we aim to identify the object(s) causing this collision sound, and output pixel-level segmentations of these objects in a frame within the collision segment. We define a collision sound as one produced when two or more objects (or parts of one object) exert force on each other as a result of an interaction. Crucially, such sound results from the interaction itself and cannot be attributed to either object alone. 
We hypothesise that collision sounds encode material and impact cues from both objects, which can be exploited to segment colliding objects.

Given this setting, our objective is to segment the colliding objects conditioned \textit{only} on the collision audio and a corresponding video frame. Our approach is trained with weak supervision, \ie we require only temporally segmented clips of collisions, without any labels of colliding objects, action semantics or object segmentation masks. 

Formally, given a collision audio $\A{\in}\R^{1 \times T}$ and an image $\I{\in}\R^{3 \times H \times W}$ from a video, we learn a function to segment object(s) responsible for the collision:
\vspace{-0.6em}
\begin{equation}
f: (\I, \A) \rightarrow \{\M_k\}_{k=1}^n,
\end{equation}
where $n{\in}\{1, 2\}$, and each $\M_k{\in}\{0, 1\}^{H{\times}W}$ denotes the mask for the $k^\text{th}$ colliding object. 
In most cases, $n{=}2$ as collisions typically involve two distinct objects.
However, we also include cases where $n{=1}$ for when two parts of the same object collide like the lid of a kettle (see Sec.~\ref{sec:dataset} for details). Due to the capabilities of current segmentation models, we leave part-based segmentation for future work.

\vspace{-0.3em}
\section{Approach}
\vspace{-0.5em}
\label{sec:approach}

We provide an overview of our approach for CS3 in Fig.~\ref{fig:approach}. Given our videos are egocentric, we use the interaction prior that sounds are often caused by the camera wearer manipulating one of the objects involved in the collision.
Guided by this hypothesis, we combine two complementary cues: (1) audio-visual correlation to localise the sound-producing object, and (2) hand-object interaction priors to identify objects held in both hands. Finally, we refine the masks and verify the colliding object(s) to obtain the corresponding segmentation masks.
In the following sections, we detail our approach. 

\vspace{-0.1em}
\subsection{Audio-conditioned Segmentation Model}
\vspace{-0.2em}
Our audio-conditioned segmentation model is trained with the weak supervision of a collision audio segment ($\A$) and its corresponding image ($\I$), to predict a segmentation mask ($\M_{AV}$), \ie $f_{av}{: }(\I, \A){\rightarrow}\M_{AV}$. We adapt a CLIP-based segmentation model \cite{cha2023learning} to interpret audio. 
To be compatible with pre-trained segmentation backbones expecting text queries, we use an audio adapter to project audio features into a text token. This allows the decoder to produce segmentation masks conditioned on the audio, reinterpreted as a text-like query. We train only the audio encoder, keeping the visual encoder-decoder frozen to leverage the generalisability of large models~\cite{luddecke2022image}. This design enables learning of audio-visual correspondences without annotations, inspired by prior sound source localisation works~\cite{park2024can, cha2023learning}.

\noindent\textbf{Audio Encoder}. 
We first process an audio spectrogram using the backbone $AE_{bkbn}$. The resulting embeddings are passed through a projection network ($AE_{prj}$) and an attentive pooling ($AE_{pool}$) layer to align them with the text token space. This produces an \emph{audio token} designed to mimic a text token in both form and semantics. We append this audio token to the tokenized prompt ``A photo of a'' and feed the sequence into the frozen CLIP text encoder ($E_{\text{CLIP}}^t$), producing an audio embedding $\A_C$. This embedding serves as a conditioning signal for the visual decoder.

\noindent\textbf{Audio-visual Decoder}. We adopt the CLIPSeg \cite{luddecke2022image} decoder, a supervised image segmentation model, for audio-driven segmentation. The input image $\I$ is processed by the CLIP visual encoder ($E_{\text{CLIP}}^v$) to extract visual features $\I_C$. The decoder is then conditioned on both $\I_C$ and the audio embedding $\A_C$ to obtain a mask, $\M_{AV} \in \R^{H \times W}$ which segments  the collision sound source.

\noindent\textbf{Training and Loss}. Following~\cite{cha2023learning}, we apply losses at both image and feature level to encourage cross-modal alignment. Given a batch of $B$ images $\I$ and corresponding audio clips $\A$, we compute masks for all image-audio combinations using the lightweight CLIPSeg decoder. This results in a tensor of masks $\M_{AV}^B{\in}\R^{B{\times}B{\times}H{\times}W}$. 

For the \textbf{image-level loss} ($\mathcal{L}_i$), we encourage similarity between audio features and segmented visual features.  We use only the positive pair masks (mask obtained from $i^{th}$ image and corresponding audio), \ie $\M{_{AV}^B}_{i,i}$, to obtain the visual features for the $i^\text{th}$ image. To ensure differentiability, we binarise the predicted mask with Gumbel-Softmax~\cite{jang2016categorical} and apply it to the input image. The resulting masked images are passed through the CLIP visual encoder to obtain segmented visual features for the $i^{th}$ sample $\I^I_i = E_{\text{CLIP}}^v(\M{_{AV}^B}_{i,i} \cdot \I_i)$. We compute cosine similarity between each image-audio pair, where for the $i^{th}$ image and $j^{th}$ audio  $S^I(i,j){=}{ (\I^I_i})^T \A_{C{_j}}$, and $^T$ is matrix transpose. To encourage alignment, we apply the InfoNCE loss:

\vspace*{-12pt}
{\small
\begin{equation}
\mathcal{L}_{i} = - \frac{1}{B} \sum_{i=1}^{B} 
\log \frac{
 S^I(i,i) / \tau 
}{
\sum\limits_{j=1}^{B} S^I(i,j) / \tau 
}  - \frac{1}{B} \sum_{i=1}^{B} 
\log \frac{
S^I(i,i) / \tau 
}{
\sum\limits_{j=1}^{B} S^I(j, i) / \tau 
}
\label{eq:eq2}
\end{equation}
}

\noindent where $\tau$ is the temperature. This loss encourages segmented image regions to align with their corresponding audio, while distinguishing them from other audio samples. 

Since the image-level loss relies only on positive pair masks, the model is not explicitly encouraged to suppress background regions when conditioned on mismatched audio. Moreover, generating localised visual features for all $B^2$ image-audio pairs is computationally expensive, as it requires a forward pass through the visual CLIP encoder ($E_{\text{CLIP}}^v$) for each pair.
To address this, we use a \textbf{feature-level loss} ($\mathcal{L}_f$) by computing visual features for all audio-visual pairs using weighted spatial pooling. 
Specifically, for each image-audio pair $(i,j)$ we mask the visual feature such as:

{\small
\begin{equation}
    \I_{i,j}^{F} = \frac{\sum_{h,w} \M{_{AV}^B}_{i,j,h,w} \cdot \I_{C_{i,h,w}}}{\sum_{h,w} \M{_{AV}^B}_{i,j,h,w}}
\end{equation}
}

\noindent i.e. we spatially pool the image features $\I_{C_{i,h,w}}$ only within the mask region $M^B_{AV_{i,j}}$. We then calculate cosine similarity, $S^F(i,j) = ({\I_{i,j}^{F}})^T \A_{C_j}$. Note that for negative pairs ($i \ne j$), $S^F(i,j)$ is computed using negative masks different from the image-level loss. We then apply the InfoNCE loss which is analogous to Eq~\ref{eq:eq2} only changing $S^I$ with $S^F$:

{\small
\begin{equation}
    \mathcal{L}_{f} = InfoNCE(S^F(i,j))
\end{equation}
}

In addition to the contrastive losses, we also incorporate an \textbf{area regularisation loss} to prevent shortcuts, \eg predicting the entire image as the mask. This constrains the area of predicted sounding regions to match a prior:

{\small
\begin{equation}
    \mathcal{L}_{r} = \sum_i \lVert \overline{M^B_{AV_{i,i}}} - p^{+}\rVert_1
    \vspace{-0.8em}
\end{equation}
}

\noindent where, $\overline{M^B_{AV_{i,i}}}$ represents predicted mask area and $p^{+}$ is a hyperparameter representing the expected region size. 

The total training loss is a weighted sum of the three components: 
\vspace{-0.8em}
\begin{equation}
    \small
    \mathcal{L} = \lambda_i\mathcal{L}_{i} + \lambda_f\mathcal{L}_{f} + \lambda_r\mathcal{L}_{r}
\end{equation}

At test time, we refine the segmentation mask ($\M_{AV}$) by cropping a region around its peak activation, and re-running the segmentation on the crop \ie $\M'_{AV}{=}f_{av}(\text{crop}(\I), \A)$.

\vspace{-0.1em}
\subsection{Hand-Object Interaction Model}
\vspace{-0.2em}
As mentioned earlier, the object held in hand is often one source of a collision sound. However, inferring it from audio alone is challenging, especially when the impact is subtle or the object is small. To capture this complementary cue, we use a Hand-Object Interaction (HOI) model~\cite{cheng2023towards} to identify objects associated with hands.
In everyday interactions, humans often use both hands to manipulate or stabilise objects, making it unclear which hand is involved in a collision. To ensure we capture the relevant object, we extract hand-associated objects from both hands. Specifically, the HOI model takes an input image $\I$ and predicts bounding boxes for any visible in-hand objects, along with the hand (left/right) association. 
Formally, let $f_{HOI}$ be the HOI model and given the input image $I$,

\vspace{-0.5 em}
\begin{equation}
{\small
    f_{HOI}(\I) \rightarrow \{obj_{LH}, obj_{RH}\}
    }
\end{equation}

where $obj_{LH} \in \R^4$ and $obj_{RH} \in \R^4$ represent the bounding box of the object in the left and right hand. 

\vspace{-0.1em}
\subsection{Collision Verification}
\vspace{-0.2em}

While the audio-conditioned segmentation and HOI model generate candidate regions, not all correspond to colliding objects. As we focus on collision sounds, spatially distant objects are unlikely responsible. We thus verify collisions with spatial proximity between object regions.

We first refine both audio-conditioned segmentation and in-hand objects using SAM2~\cite{ravi2024sam2}, which we denote as $f_{SAM}$. To obtain segmentation masks for the in-hand objects, we prompt SAM2 with the bounding boxes $obj_{LH}$ and $obj_{RH}$: $(\M_{left}, \M_{right}){=}(f_{SAM}(obj_{LH}), f_{SAM}(obj_{RH}))$. 
For the audio-conditioned segmentation, we extract a bounding box around the peak activation in the refined mask $\M'_{AV}$ and apply SAM2: $\M_{av}{=}f_{SAM}(bbox(\M'_{AV}))$.

To verify which objects collide, we first merge any pair of masks with an Intersection-over-Union (IoU) above a threshold $\alpha$. As collision sounds imply physical contact, we compute the pairwise distance between the three candidate masks: $\M_{av}, \M_{left}, \M_{right}$ and select the two closest as the colliding pair. 
If no distance falls below a threshold $\beta$, we assume a single-object collision and prioritize masks in the order: $\M_{right}$,  $\M_{left}$, $\M_{av}$. %

\vspace{-0.3em}
\section{Collision Sound Source Datasets}
\vspace{-0.2em}
\label{sec:dataset}

Since there are no existing datasets suitable for our CS3 task, we curate our own: EPIC-CS3 and Ego4D-CS3. 

\noindent\textbf{Datasets.} We use videos from EPIC-Kitchens-100 \cite{damen2018scaling} and Ego4D \cite{Grauman_2022_CVPR}, which feature diverse egocentric object interactions and collision sounds. EPIC-Kitchens captures 100 hours of unscripted cooking activities over 45 home kitchens. Ego4D is larger, with over 3,600 hours of daily activities ranging from baking and sports to sewing and gardening. As neither dataset include collision sound segmentations, we create these ourselves.  

\noindent\textbf{Extracting Action Sounds.} We first extract short audio-visual clips containing collision sounds. For EPIC-Kitchens, we use EPIC-Sounds annotations~\cite{huh2023epic}, which provide start and end times for sound events. 
For Ego4D, which lacks sound annotations, we follow~\cite{chen2024action2sound} and use narrations timestamps, extracting a 3 second clip centred around each. We filter out clips labelled as a social scenarios, dominated by speech/music or with low average amplitude.

\begin{figure}[t]
{\includegraphics[width=\linewidth]{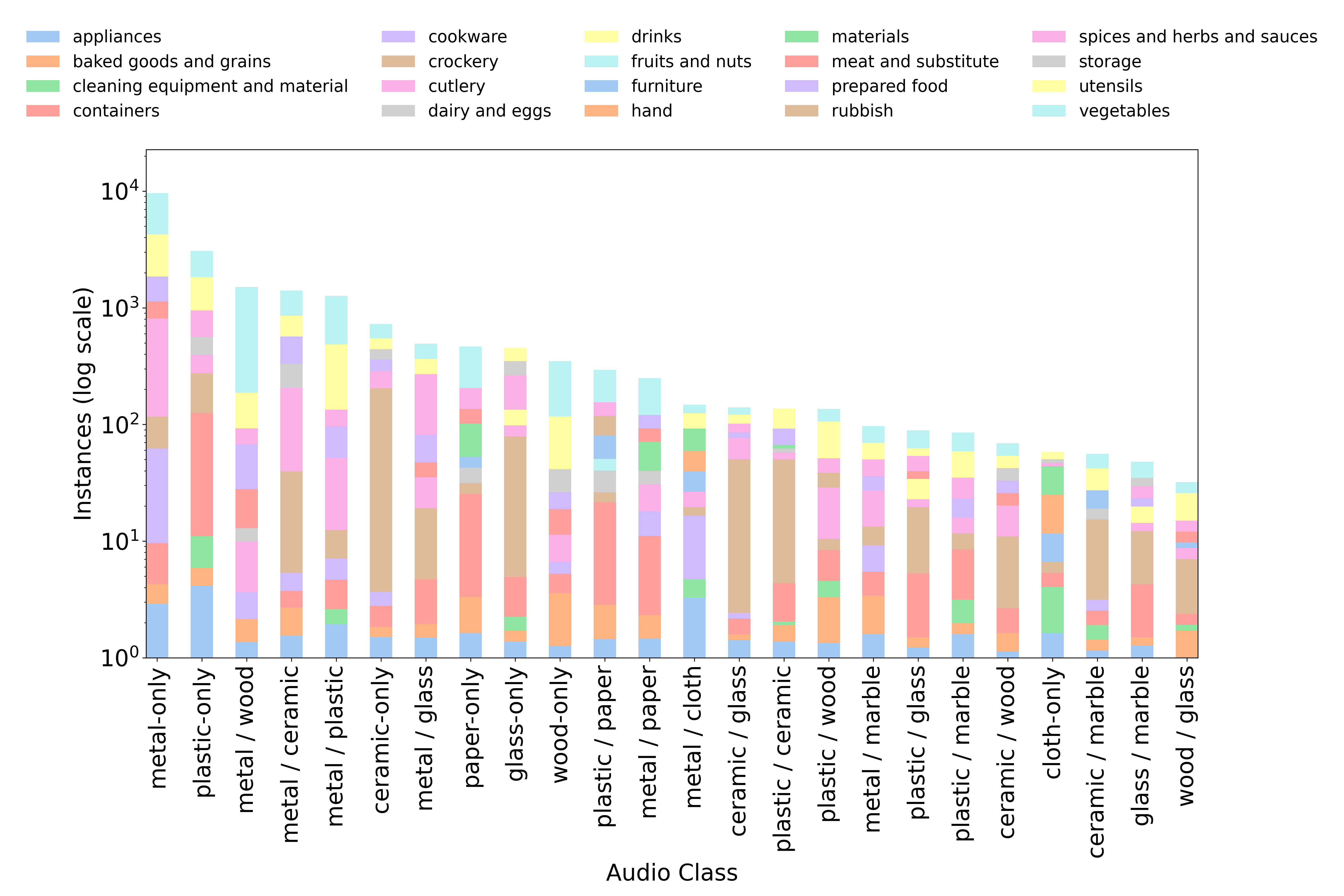} }
\vspace{-2em}
   \caption{\textbf{Distribution of EPIC-CS3} over the predicted sound class and noun category in the corresponding action.}
   \vspace{-0.5em}
    \label{epic_classes_distribution}
\end{figure}

\noindent\textbf{Identifying Collisions.} Not all action sounds are collisions, thus we further filter clips to retain only those with collision sounds. For EPIC-Kitchens, we select 24 EPIC-Sounds~\cite{huh2023epic} categories typically caused by collisions (\eg `plastic/paper', `metal/glass', `wood-only'), resulting in 20,966 clips. For Ego4D, we classify the audio of each clip using publicly available Auditory SlowFast \cite{Kazakos2021SlowFastAuditory} model trained on EPIC-Sounds, and retain the clips classified into the same 24 classes above. This results in 75,704 clips predicted as collision-related. 

\noindent\textbf{Masks of Colliding Objects.} To evaluate CS3, we construct clean, annotated test sets with ground-truth masks of the colliding objects. The training set remains unannotated as our method is weakly supervised. Note that while most collisions involve two distinct objects, some involve only a single object with interacting parts, \eg a kettle and its base or a bottle and its lid. 

We follow the original train/test splits of each dataset. For EPIC-Kitchens we rely on VISOR \cite{darkhalil2022epic} to obtain ground-truth masks. We manually select a frame within each clip where the colliding objects are visible and segmented in VISOR. %
If none exists, we use AOT \cite{yang2021aot} to propagate the nearest available masks to a selected frame. Masks are manually reviewed and corrected using Toras~\cite{torontoannotsuite}. Clips without a corresponding VISOR mask are discarded, resulting in 614 annotated test samples. 

For Ego4D, we manually verify each collision clip, discarding those where the collision is out-of-view. We then select a frame where the collision is clearly visible. Each selected frame is annotated with bounding boxes for the colliding objects, which are used to prompt SAM2 \cite{ravi2024sam2} for mask generation. As with EPIC, all outputs are manually reviewed and corrected with Toras \cite{torontoannotsuite}. 
This yields a test set of 528 annotated samples.
\begin{figure}[t]
    {\includegraphics[width=0.95\linewidth]{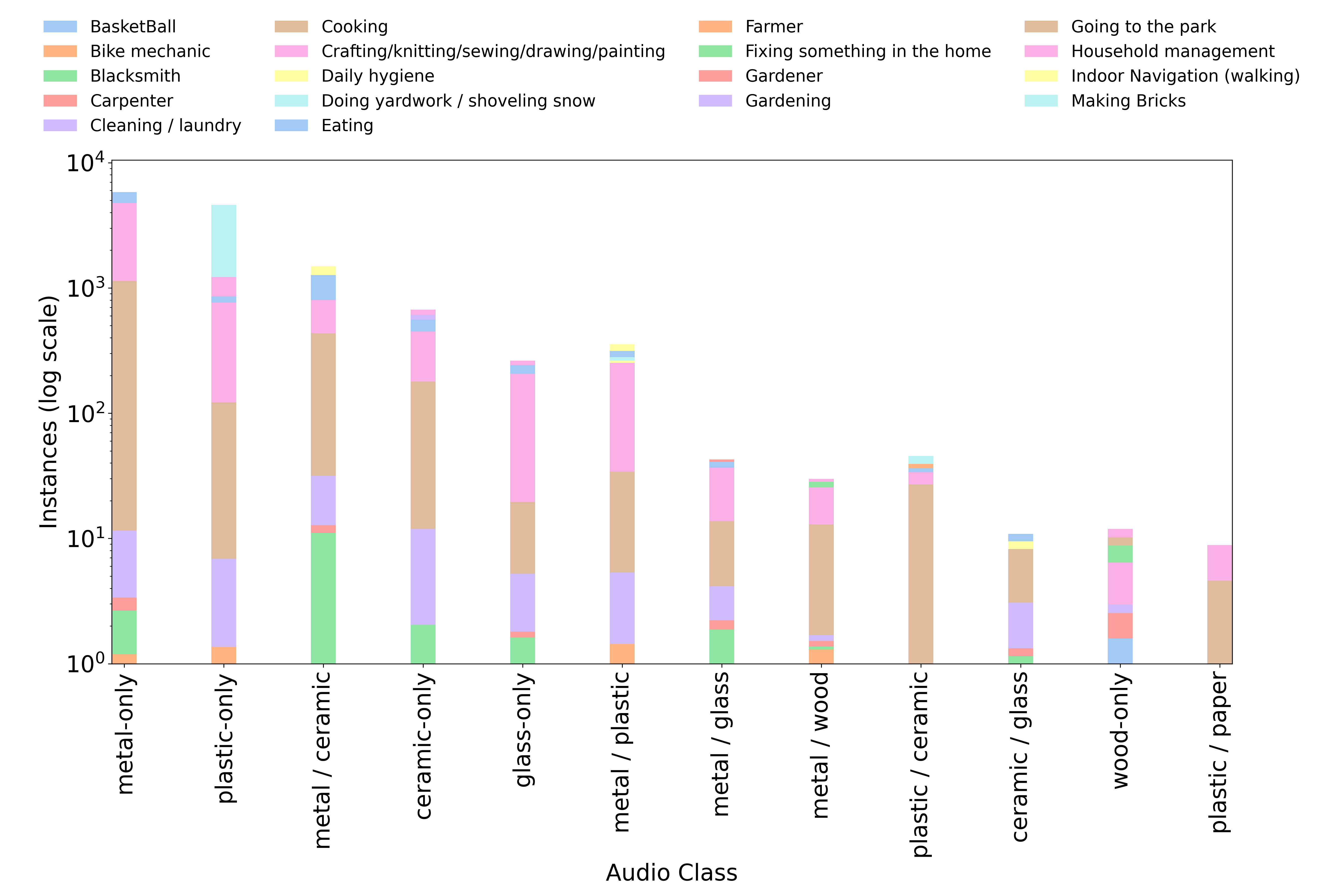} }    
    \vspace{-1em}
    \caption{\textbf{Distribution of Ego4D-CS3} over the predicted sound classes and the scenario causing the sound.}
    \vspace{-1em}
    \label{ego4d_classes_distribution}
\end{figure}

\noindent\textbf{Dataset Summary. } We produce two datasets: EPIC-CS3 and Ego4D-CS3 which we will release publicly with code. EPIC-CS3 contains 20,352 train and 614 test clips.
In EPIC-CS3 76.7\% of test collisions involve two objects and 23.3\% are single object. Analogously, Ego4D-CS3 has 79,148 train and 528 test samples, with 87.7\% involving two objects and 12.3\% showcasing a single object collision.

\begin{figure}[t]
    {\includegraphics[width=\linewidth]{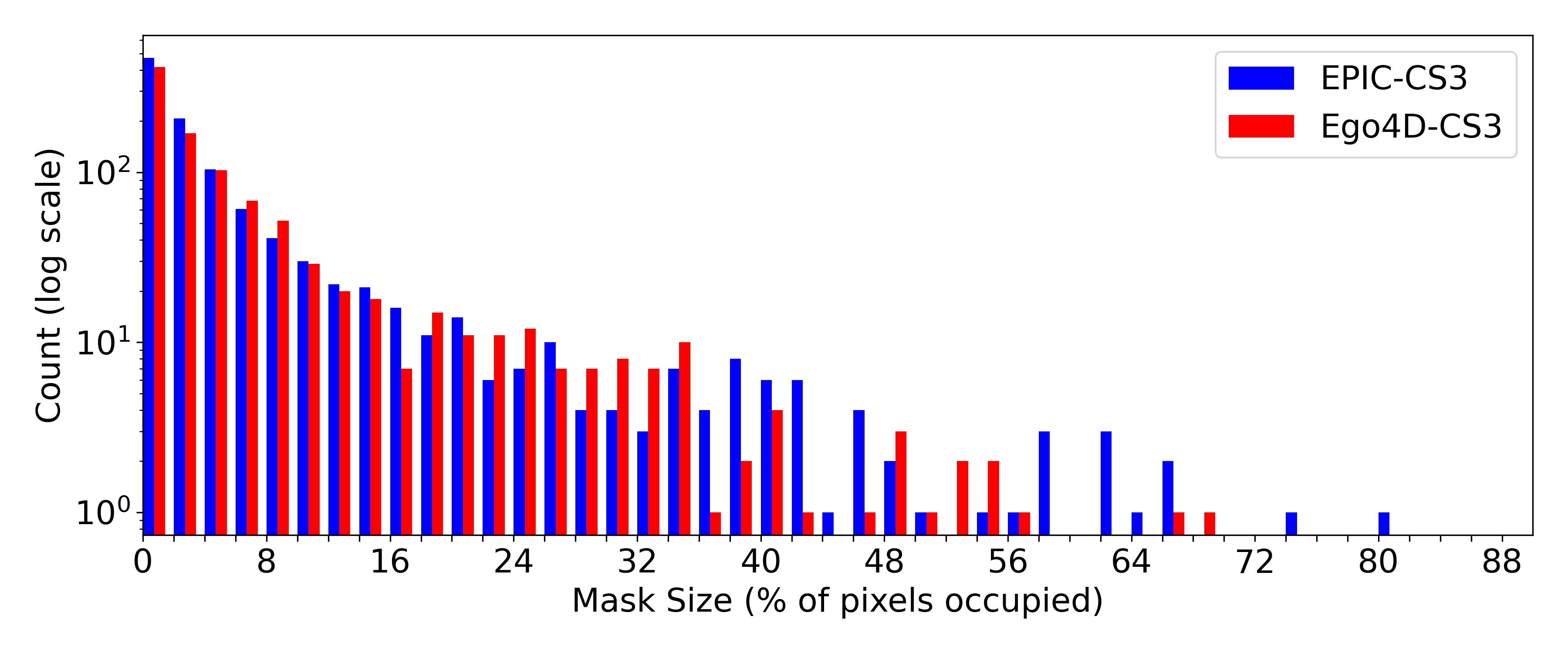} }    
    \vspace{-2.5em}
    \caption{\textbf{Distribution of Masks Sizes} by percentage of pixels occupied. Many small objects make segmentation challenging.}
    \vspace{-0.5em}
    \label{mask_sizes_distribution}
\end{figure}

To gain insight into our datasets, we analyse the distribution of predicted collision sound classes and associated contexts.  Fig.~\ref{epic_classes_distribution} shows EPIC-CS3 by predicted sound class and noun category (from the corresponding action). Fig.~\ref{ego4d_classes_distribution} shows Ego4D-CS3 by sound class and scenario.
Both datasets exhibit a long-tailed distribution of collision types, involving different materials. 
These collisions span diverse scenarios and object types, adding to the task's complexity. 
We also analyse the distribution of object sizes (Fig. \ref{mask_sizes_distribution}) 
and find that most colliding objects occupy $<$10\% of the of the frame which further increases the challenge of our task.

\vspace{-0.2em}
\section{Experiments}
\vspace{-0.3em}
\label{sec:experiments}

\vspace{-0.1em}
\subsection{Implementation Details}
\vspace{-0.2em}
We use ViT-B/16 CLIP~\cite{radford2021learning} image and text encoders, with CLIPSeg~\cite{luddecke2022image} for visual grounding and~\cite{cheng2023towards} as the HOI model. Segmentation masks from bounding boxes are generated with SAM2~\cite{ravi2024sam2} and BEATS~\cite{chen2022beats}, pretrained on AudioSet~\cite{gemmeke2017audio}, serves as the audio backbone ($AE_{bkbn}$). 
During training, we finetune the audio backbone and train the projection ($AE_{prj}$) from scratch for $50,000$ steps with batch size $32$ and learning rate $1e{-}6$, on $4$ NVIDIA H100 GPUs. We set $\lambda_i{=}1$, $\lambda_f{=}1$ and $p^+{=}0.1$ for both the datasets and use $\lambda_r{=}1$ for EPIC-CS3 and $\lambda_r{=}0.1$ for Ego4D-CS3. For collision verification we set $\alpha{=}0.6$ and $\beta{=}15$.
We train using $2$-second audio clips sampled at $16$ kHz and randomly sample a frame within the collision segment as the paired image of size $352\times352$.
For Ego4D-CS3, background noise often dominates. Thus, we identify the peak amplitude and randomly select a $[0.5,1.5]$ second window around this peak. A random frame from this window is used as the image. At test time, we use the fixed annotated frame. 

\vspace{-0.1em}
\subsection{Evaluation Metrics}
\vspace{-0.2em}
Unlike prior sound source localisation works~\cite{zhou2022audio, hu2022mix}, which either predict a single mask or exactly two masks, our task allows for a variable number of colliding objects ($1$ or $2$). Thus, our evaluation metric should assess both segmentation quality and predicted number of masks. We evaluate performance using mean Intersection-over-Union (mIoU) and area under the cuve (AUC). Following prior multi-object segmentation works \cite{karazija2022unsupervised, kipf2021conditional}, we use Hungarian matching to associate  predicted and ground-truth masks. 
For AUC, we follow ~\cite{park2024can, chen2021localizing}, computing the fraction of predicted masks with IoU above thresholds from $0$ to $1$ in steps of $0.05$, which are then averaged. %

\begin{table}
    \centering
 \resizebox{\linewidth}{!}{
    \begin{tabular}{lccccccc}
    \toprule
    & \textbf{Two} & & \multicolumn{2}{c}{\textbf{EPIC-CS3}}& & \multicolumn{2}{c}{\textbf{Ego4D-CS3}} \\
    \cmidrule(lr){3-5} \cmidrule(lr){6-8}
    \textbf{Method} & \textbf{Masks}& & \textbf{mIoU} $\uparrow$ & \textbf{AUC} $\uparrow$ & &\textbf{mIoU} $\uparrow$ & \textbf{AUC} $\uparrow$ \\
    
    \midrule
    Random & \xmark & &2.16 & 3.96 && 5.81 & 6.42 \\
    Centre & \xmark && 6.95 & 7.43 && 6.06 & 6.82 \\
    CLIPSound \cite{park2024can} & \xmark && 6.46 & 6.97 && 7.53 & 8.25 \\
    Mix-Localize \cite{hu2022mix} & \cmark  && 7.20 & 7.53 && 7.17 & 7.65 \\
    Ours (AV) & \xmark && 13.13 & 13.48 && 8.16 & 8.48\\
    Ours & \cmark && \textbf{39.38} & \textbf{39.60} && \textbf{38.34} & \textbf{38.43} \\
    
    \bottomrule
    \end{tabular}
     }
     \vspace{-1 em}
    \caption{\textbf{Comparison with existing methods}, both one- and two-mask. Our full model significantly outperforms all baselines. %
    }
    \label{tab:sota}
    \vspace{-1 em}
\end{table}

\subsection{Quantitative Comparison}
\vspace{-0.2em}
We report quantitative results on EPIC-CS3 and Ego4D-CS3 in Table~\ref{tab:sota}, comparing our method against naive heuristics and prior sound source localisation methods.
\textbf{Random} uses the audio-conditioned segmentation model with the pre-trained components (CLIP visual encoder, CLIPSeg decoder, BEATS audio encoder), but random weights for the projection network which produces the audio token. 
\textbf{Centre} predicts a fixed square region occupying $10\%$ of the image centre. This is evaluated to measure the bias in egocentric videos as active objects can often be present at the centre of the image. 
\textbf{CLIPSound \cite{park2024can}} is a recent  sound source localisation method trained on VGGSound \cite{chen2020vggsound}. These three baselines produce a single mask per image. 
For samples with two colliding objects, the unmatched ground-truth mask is treated as empty during evaluation.  
We also compare against \textbf{Mix-Localize}~\cite{hu2022mix}, a recent multi-sound source localisation method. %
Since this approach trains with synthetic mixtures from individual sources we use the released weights. 
Finally, we include a variant of our method \textbf{Ours~(AV)}, which uses only the audio-conditioned segmentation model (without hand-object or collision reasoning) and thus predicts a single object mask. %

As expected, Random yields the lowest performance, highlighting the difficultly of the task.  %
Despite the centre bias present in egocentric video, Centre also performs poorly.
Interestingly, CLIPSound~\cite{park2024can} performs comparably to Center on EPIC-CS3 (mIoU: 6.95 vs. 6.46) and only slightly better on Ego4D-CS3 (mIoU:  6.06 vs.7.53).
Mix-Localize~\cite{hu2022mix} also performs poorly on our task (mIoU: $7.20$ on EPIC-CS3, $7.17$ on Ego4D-CS3), %
indicating that general-purpose sound-source localisation methods do not transfer well to collision sound source segmentation. %
Our audio-conditioned backbone (Ours (AV)) improves over all baselines, even by predicting a single mask. Our full model achieves the best results, reaching mIoU and AUC of $39.38$ and $39.60$ on EPIC-CS3, and $38.34$ and $38.43$ on Ego4D-CS3, at least $4.7\times$ better than the best baseline on AUC.

\vspace{-0.1em}
\subsection{Ablations}
\vspace{-0.2em}
We ablate different design choices with EPIC-CS3.
\begin{table}
    \centering
 \resizebox{0.8\linewidth}{!}{
    \begin{tabular}{lcc}
    \toprule
    \textbf{Method} & \textbf{mIoU} $\uparrow$ & \textbf{AUC} $\uparrow$ \\
    
    \midrule
    Ours & 39.38 & 39.60 \\
    Ours w/o SAM(AV) & 37.92 & 37.93 \\
    Ours w/o SAM(AV) w/o crop & 37.67 & 37.75 \\
    Ours w/o SAM(AV) w/o crop w/o HOI & 10.01 & 10.11 \\
    \bottomrule
    \end{tabular}
     }
    \vspace{-0.8em}
    \caption{\textbf{Ablation}. Performance drops as SAM refinement, cropping, and hand-object reasoning are removed, highlighting the importance of each for accurate collision sound source segmentation.
    }
    \label{tab:ablation_two_step}
    \vspace{-1 em}
\end{table}
\noindent \textbf{Method Ablation.} We ablate  our method in Table~\ref{tab:ablation_two_step}. Removing SAM2 refinement for the audio-conditioned segmentation mask (\textbf{Ours w/o SAM (AV)}) drops performance in both mIoU and AUC from $39.38$ and $39.60$  to $37.92$ and $37.93$. Removing the cropping step used to refine the audio-conditioned segmentation further reduces performance ($37.67$ mIoU and $37.75$ AUC). Finally, removing the HOI model and relying solely on audio-conditioned segmentation leads to a major drop ($10.01$ mIoU and $10.11$ AUC), highlighting the importance of the hand-object cues.

\begin{table}
    \centering
    \setlength{\tabcolsep}{10pt}
 \resizebox{0.6\linewidth}{!}{
    \begin{tabular}{lccc}
    \toprule
    \textbf{Method} & \textbf{mIoU} $\uparrow$ & \textbf{AUC} $\uparrow$ \\
    
    \midrule
    HOI (Right) & 25.66 & 26.48 \\
    HOI (Left) & 19.00 & 20.25 \\
    HOI (Right-AV) & 31.28 & 31.64 \\
    HOI (Left-AV) & 23.82 & 24.57 \\
    HOI (Touch) & 34.03 & 34.80 \\
    HOI (Right-Left)  & 37.63 & 37.81 \\
    Ours &  \textbf{39.38} & \textbf{39.60} \\
    \midrule
     \color{gray} Oracle & \color{gray} 44.10 & \color{gray} 44.38 \\
    \bottomrule
    \end{tabular}
     }
     \vspace{-1em}
    \caption{\textbf{Hands.} Comparing different hand-based strategies and their integration with audio confirms that hands offer a strong prior, with audio offering complementary cues.}
    \label{tab:ablation_hands}
    \vspace{-1 em}
\end{table}
\noindent \textbf{Hands: How and Which?} As hand-held objects are central to our approach, we investigate different strategies for selecting colliding objects using the HOI model in Table~\ref{tab:ablation_hands}. HOI (Right) and HOI (Left) use only the object in the right or left hand. HOI (Right-Light) assumes a two-object collision between objects in both hands. Similarly, HOI (Right-AV) and HOI (Left-AV) pair one hand-held object with the audio-conditioned segmentation. HOI (Touch) uses the right-hand object and the object predicted to be in contact with it~\cite{cheng2023towards}. Finally, \textbf{Oracle} uses ground-truth to pick the best matching candidate masks based on IoU. 

HOI (Right) outperforms HOI (Left), suggesting the right-hand object is more frequently involved in collisions. Incorporating the audio-conditioned segmentation improves performance by $\sim$24\% (Right-AV) and $\sim$25\% (Left-AV),  indicating that audio provides complementary cues. Using both hands (Right-Left) yields better performance than individual or touch-based variants, reinforcing the strong hand bias in egocentric interactions consistent with recent findings~\cite{chen2024unveiling}. Our full model, which combines hand-object and audio-conditioned cues, achieves the best performance. The Oracle adds a further 4.7 mIoU improvement over using only the left and right-hand objects, suggesting additional untapped potential in the selection of the right set of masks.

\vspace{-0.1em}
\subsection{Analysis}
\vspace{-0.2em}
Using EPIC-CS3, we analyse the impact of three factors: 

\begin{figure}
    \centering
    \includegraphics[width=0.98\linewidth]{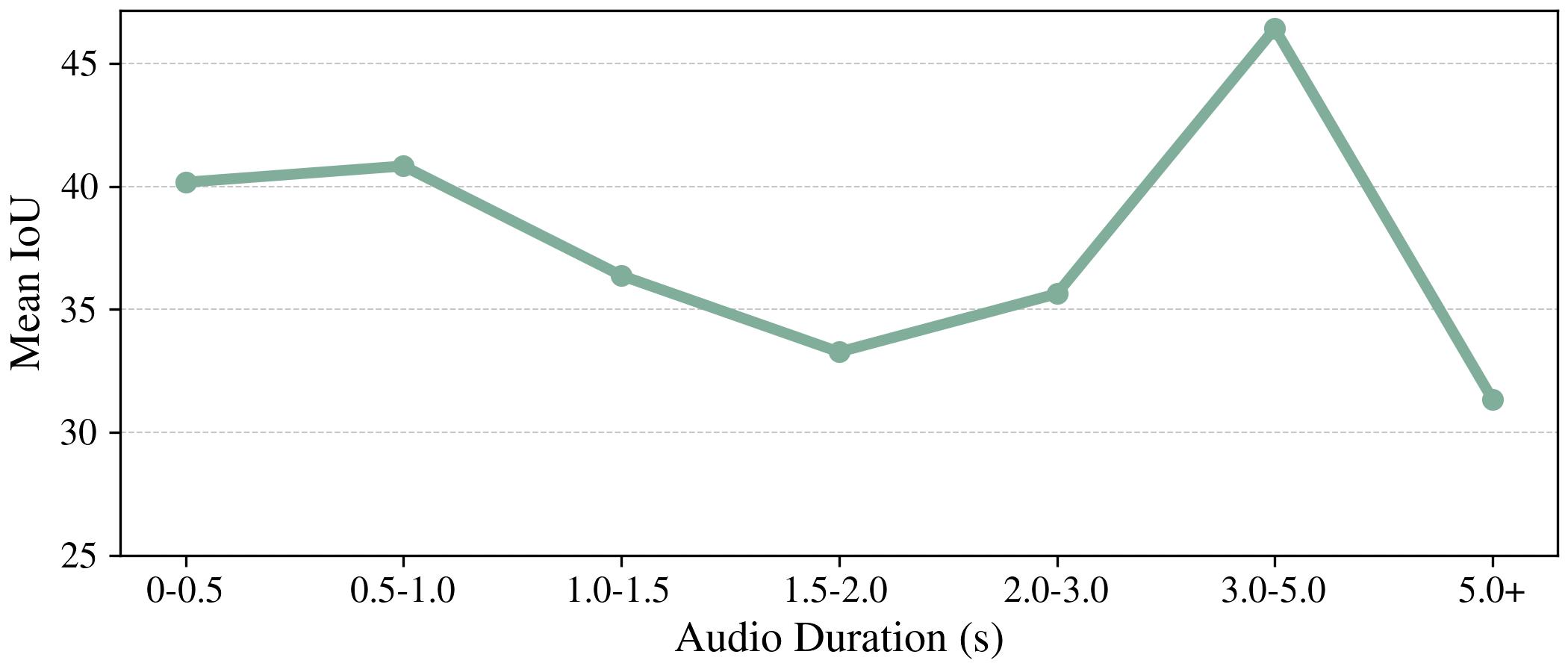}
    \vspace{-0.8em}
    \caption{\textbf{Performance by audio duration}. Performance drops for clips ${>}$5 seconds, likely due to noisy or irrelevant audio.}
    \label{fig:analysis_audio_dur}
    \vspace{-1 em}
\end{figure}

\begin{figure}
    \centering
    \includegraphics[width=0.98\linewidth]{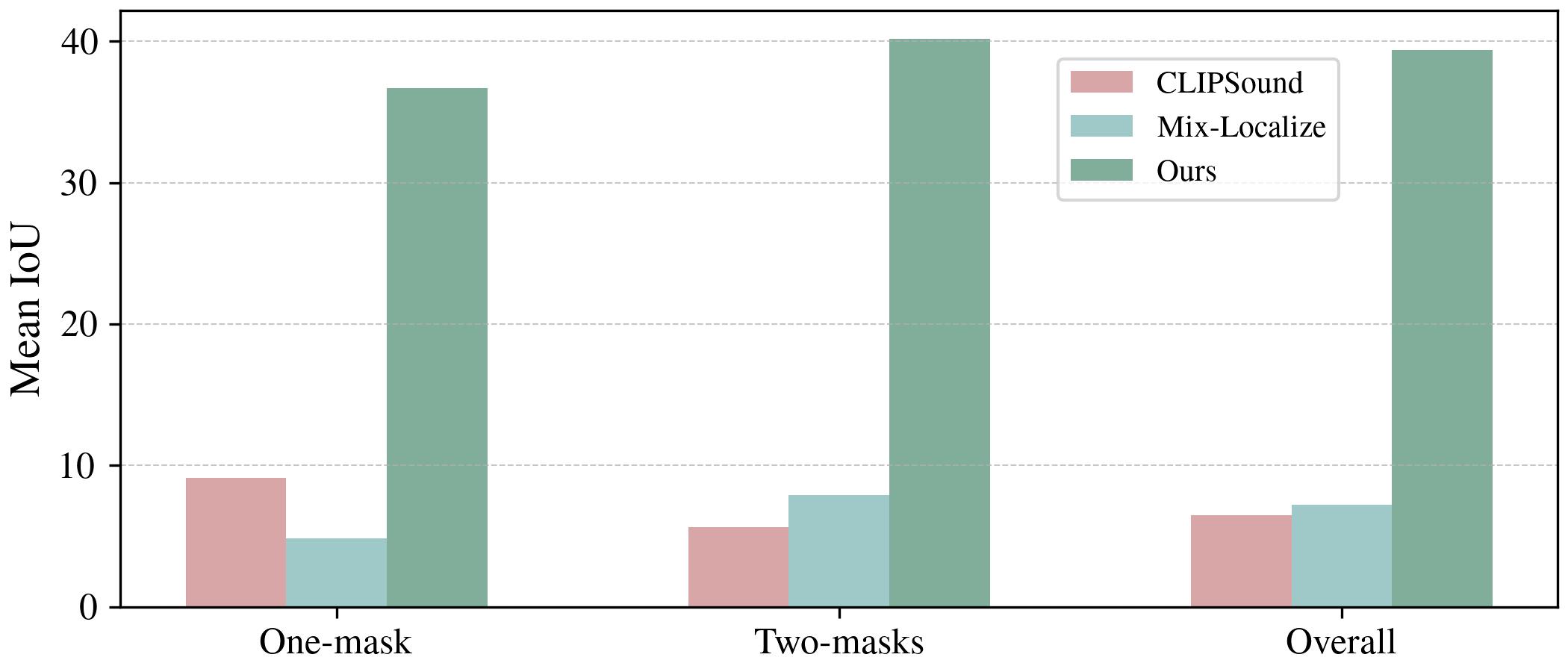}
    \vspace{-1.2 em}
    \caption{\textbf{Performance by number of masks}. Our approach performs strongly in both single and two-mask cases.}
    \label{fig:analysis_n_masks}
    \vspace{-1 em}
\end{figure}
\noindent\textbf{Performance by Audio Duration.} Test clip durations range from $0.2$ to $13$ seconds. To assess the impact of duration, we evaluate performance across different audio lengths in Fig.~\ref{fig:analysis_audio_dur}. For short clips ($<1$s), which make up $\sim$65\% of the test set, performance is $\sim$40\%. The highest performance ($\sim$45 mIoU) occurs for audio durations between $3–5$ seconds. Beyond $5$ seconds, performance drops marginally, suggesting that longer clips often include irrelevant or noisy background audio that hinders segmentation.

\begin{figure}
    \centering
    \includegraphics[width=0.98\linewidth]{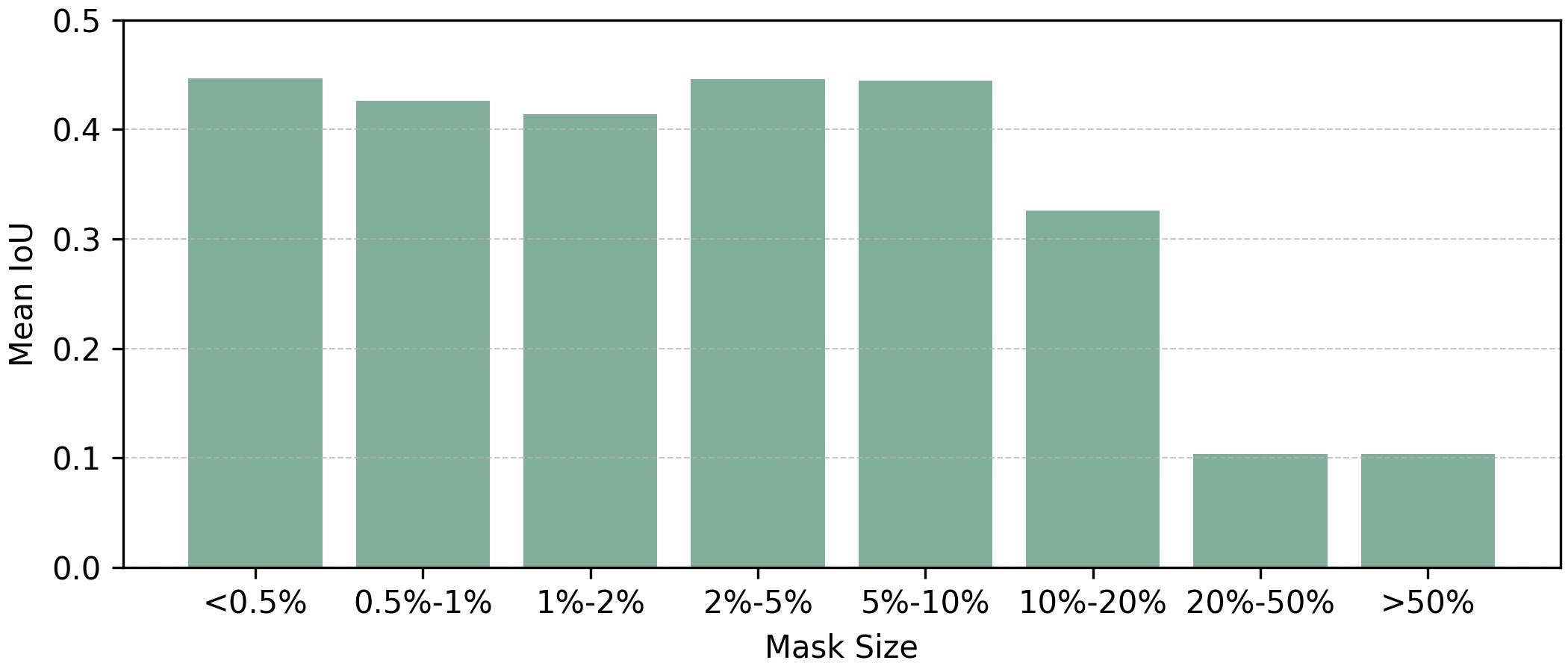}
    \vspace{-0.8em}
    \caption{\textbf{Performance by Object Size}. Our approach successfully tackles the challenge of small objects.}
    \label{fig:mask_size_analysis}
    \vspace{-1.2 em}
\end{figure}

\begin{figure}
    \centering
    \includegraphics[width=\linewidth]{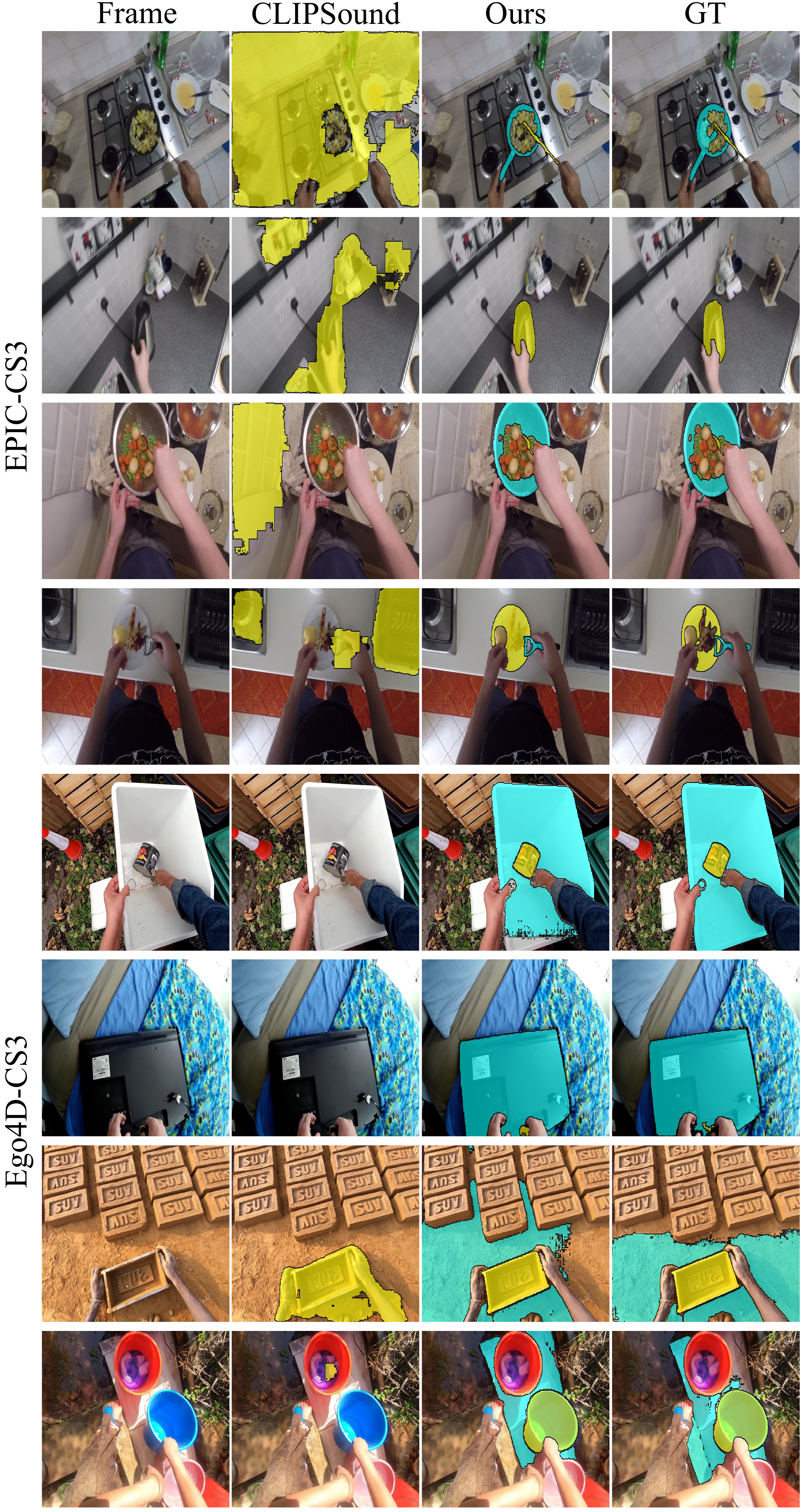}
    \vspace{-2em}
    \caption{\textbf{Qualitative Results.} Our model segments sound-producing regions more accurately than prior sound source segmentation method CLIPSound, especially  when objects are small. }
    \vspace{-1.5em}
    \label{successful_samples}
\end{figure}

\begin{figure}
    \centering
    \includegraphics[width=\linewidth]{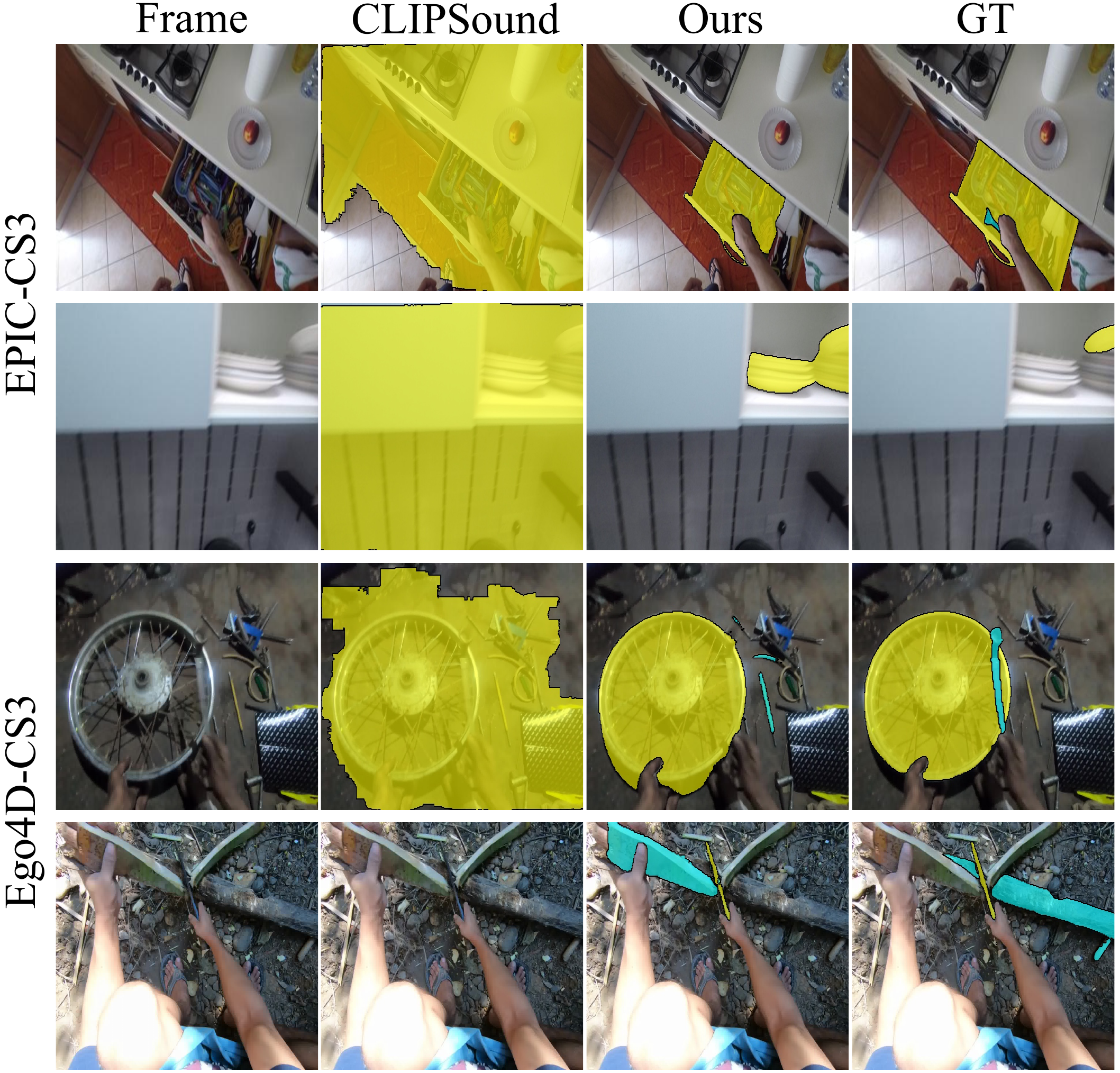}
    \vspace{-2em}
    \caption{\textbf{Failure Cases.} Our approach struggles in scenes with similar-material objects, where the sound-source is  ambiguous. }
    \vspace{-1.5em}
    \label{failure_samples}
\end{figure}

\noindent\textbf{Performance by Number of Masks}. We evaluate our approach on two subsets: collisions involving a single object (one mask), and collisions with two distinct objects (two masks).
We compare against CLIPSound~\cite{park2024can}, which always predicts one mask, and Mix-Localize \cite{hu2022mix}, which always predicts two. Our method outperforms both baselines by a large margin, achieving a $4$x mIoU for the single-mask subset and $5$x for the two mask subset. Our performance is strong for both cases, but is slightly higher for the two-object collisions ($40.205$ \vs $36.678$).

\noindent\textbf{Performance by Object Size.} Unlike traditional sound source localisation datasets, our data includes significantly smaller objects. In EPIC-CS3, the average object occupies only $10\%$ of the image, compared to $40\%$ in VGG-SS~\cite{chen2021localizing}. To assess the impact of object size, we analyse performance by mask area in Fig.~\ref{fig:mask_size_analysis}.
Despite the challenge, our model performs well even on very small objects: achieving $44.69$ mIoU for masks under 0.5\% of the image. Performance drops for large objects ($>20\%$), possibly due to a bias towards the dataset mean size ($10\%$).

\vspace{-0.1em}
\subsection{Qualitative Results}
\vspace{-0.2em}

Fig.~\ref{successful_samples} and~\ref{failure_samples} present qualitative examples of our approach in comparison to CLIPSound~\cite{park2024can}.
Successful examples (Fig.~\ref{successful_samples}) highlight our model's ability to accurately segment small and partially occluded objects (rows 1, 3, 4, 6), distinguish single-object collisions (row 2) and handle cluttered scenes with distractors (rows 7, 8). 
In contrast, CLIPSound predicts a single, often large, mask failing to capture the complexity of the interactions. Failure cases (Fig. \ref{failure_samples}) include very crowded scenes (the cutlery drawer in row 1), and when there are multiple objects of the same material which could produce the same sound (metal tools in row 3 and the wooden logs in row 4). 

\vspace{-0.7em}
\section{Conclusion}
\vspace{-0.3em}
\label{sec:conclusions}

In this paper, we introduced the task of Collision Sound Source Segmentation (CS3) where the goal is to identify objects responsible for collision sounds in visually complex, egocentric scenes. We proposed a weakly-supervised method that combines audio-conditioned segmentation with egocentric cues from hand-object interactions, and introduced two benchmark datasets for evaluation. Our approach significantly outperforms prior methods. Future work may involve identifying the object parts involved in a collision.

\noindent \textbf{Acknowledgements: } This project is supported by EPSRC Program Grant Visual AI (EP/T028572/1). O Emara is supported by UKRI CD in Interactive AI (EP/S022937/1). H Doughty is supported  by the Dutch Research Council~(NWO) under a Veni grant (VI.Veni.222.160). We acknowledge the usage of EPSRC Tier-2 Jade clusters for initial experiments. The authors also acknowledge the use of Isambard-AI National AI Research Resource (AIRR). Isambard-AI is operated by the University of Bristol and is funded by the UK Government’s Department for Science, Innovation and Technology (DSIT) via UK Research and Innovation; and the Science and Technology Facilities Council [ST/AIRR/I-A-I/1023]. We also extend our gratitude to SURF (\url{www.surf.nl}) for granting compute resources from the National Supercomputer Snellius.

{\small
\bibliographystyle{ieee_fullname}
\bibliography{main}
}

\end{document}